\documentclass[a4paper]{article}

\usepackage{INTERSPEECH2019}
\usepackage{url}

\title{Multi-lingual Dialogue Act Recognition with Deep Learning Methods}
\name{Ji\v{r}\'i Mart\'inek$^{1, 2}$, Pavel Kr\'al$^{1, 2}$, Ladislav Lenc$^{1, 2}$, Christophe Cerisara$^3$}

\address{
        $^1$Dept. of Computer Science \& Engineering\\
        University of West Bohemia\\
        Plze\v{n}, Czech Republic\\
        $^2$NTIS - New Technologies for the Information Society\\
        University of West Bohemia\\
        Plze\v{n}, Czech Republic\\
        $^3$LORIA UMR 7503\\
        BP 239 - 54506 Vandoeuvre, FRANCE}
\email{\{jimar,pkral,llenc\}@kiv.zcu.cz,cerisara@loria.fr}

\begin{document}

\maketitle
\begin{abstract}
This paper deals with multi-lingual dialogue act (DA) recognition. 
The proposed approaches are based on deep neural networks and use word2vec embeddings for word representation.
Two multi-lingual models are proposed for this task. 
The first approach uses one general model trained on the embeddings from all available languages.
The second method trains the model on a single pivot language and a linear transformation method is used to project other languages onto the pivot language.
The popular convolutional neural network and LSTM architectures with different set-ups are used as classifiers.
To the best of our knowledge this is the first attempt at multi-lingual DA recognition using neural networks. 
The multi-lingual models are validated experimentally on two languages from the Verbmobil corpus. 
\end{abstract}

\section{Introduction}
Nowadays, the importance of multi-lingual and cross-lingual natural language processing (NLP) methods is still growing.
Another important research direction is the usage of deep neural networks that learn parameters implicitly and do not require manual feature engineering.
Both research directions respectively help to significantly reduce the amount of human annotation efforts and improve the applicability of the models to various corpora and contexts.
Many researchers have proposed multi-lingual approaches based on neural networks for a wide spectrum of NLP tasks, including document classification~\cite{klementiev2012inducing}, named entity recognition~\cite{agerri2016robust} and semantic role labelling~\cite{bjorkelund2009multilingual}.
Unfortunately, to the best of our knowledge, research in the multi-lingual automatic DA recognition field is scarce.

DA recognition is an important step in dialogue understanding and it plays a pivotal role in dialogue management~\cite{jekat1995dialogue}.
Any improvement in this task may increase the performance of the whole dialogue system.
In this paper we propose and compare several methods for multi-lingual and cross-lingual DA recognition.
The methods utilize deep neural networks and word2vec embeddings are used for word representation.
The first approach trains one general model on annotated DAs from all available languages.
This model is thus able to perform DA recognition in multiple languages simultaneously.
The second method trains the model only on one language and cross-linguality is achieved by a linear semantic space transformation.

We employ two standard neural network topologies with different set-ups, namely the convolutional neural network (CNN) and the long short-term memory (LSTM), and we compare and evaluate them on the Verbmobil corpus~\cite{alexandersson1998dialogue}.
Implementations of all presented methods are publicly available for research purposes. 

\section{Related Work}
Traditional DA recognition methods usually create complex handcrafted features using one of, or a combination of the following types of information:

\begin{itemize}
\item lexical and syntactic information
\item prosodic information
\item dialogue history
\end{itemize}

Several lexical models are proposed including Bayesian approaches such as n-gram language models~\cite{reithinger1997dialogue,stolcke2000dialogue,ang2005automatic} and also non-Bayesian methods -- semantic classification trees~\cite{mast1995automatic}, transformation-based learning (TBL)~\cite{samuel1998dialogue} or memory-based learning~\cite{Rotaru02}. 
Syntactic features that are created using a full parse tree are considered for instance in~\cite{kral2014automatic}.
Prosodic information is often used to provide additional clues to recognize DAs as presented for instance in~\cite{Shriberg98}.
Dialogue history (the sequence of DAs) can been used to predict the most probable next dialogue act and it can be modeled for example by
hidden Markov models~\cite{stolcke2000dialogue} or Bayesian networks~\cite{Keizer02}.

Nowadays, DA recognition is often realized with simple word features and sophisticated neural models including popular CNNs and an LSTM.
The features are represented by word embeddings provided by word2vec~\cite{mikolov2013efficient} or glove~\cite{pennington2014glove}.
Both word2vec and glove are used in~\cite{N16-1062} where a CNN (or an LSTM) is used to create a vector representing a DA.
This vector is then fed into a feed forward network utilized for classification.
The experiments on several DA corpora have shown that the CNN performs slightly better than the LSTM in this case.
Another approach using only raw word forms as an input is presented in~\cite{cerisara2018effects}.
The LSTM with word2vec embeddings is used to model the DA structure as well as for DA prediction with very good results.
A similar LSTM model with word2vec embeddings has been also proposed for DA recognition in~\cite{khanpour2016dialogue}.
The authors experimented with different embedding sizes and network hyper-parameters.
Duran et al.~\cite{duran2018probabilistic} propose new features called "probabilistic word embeddings" which are based on word distribution across DA-set.
The experiments show that these features perform slightly better than word2vec.

The above mentioned approaches are mainly evaluated on English language using Switchboard (SwDA)~\cite{jurafskyswitchboard} or Meeting Recorder Dialog Act (MRDA)~\cite{shriberg2004icsi} corpora.
Some methods are tested on Spanish (see DIHANA~\cite{benedi2006design} corpus), Czech~\cite{Kral05a}, French~\cite{barahona2012building} or on German (see Verbmobil~\cite{jekat1995dialogue} corpus) languages.

These corpora are annotated according to different annotation schemes and they contain different DA labels, although relevant efforts towards a standardization of DA annotations have been made~\cite{ISO}.
Their direct usage for multi-lingual DA recognition is difficult, because a mapping between the different tag-sets is required.
To the best of our knowledge, only the Verbmobil corpus contains a large enough number of annotated dialogues in multiple languages.
Mapping DA annotations that are both satisfactory from the linguistic point of view and easily usable computationally is challenging and complex, therefore we have thus chosen the robust and well-known Verbmobil corpus in our experiments.

\section{Multi-lingual DA Recognition}
\label{sec:lin_transf}
This section starts by describing the two methods we use to achieve multi-linguality.
The neural network architectures are described next. 

\subsection{Multi-lingual Model}
\label{sec:mlm}
Let $\mathbb{L} = \{L_1,L2,...,L_M\}$ be a set of languages with available annotated DAs and $\mathbb{T}_{L_i}$ be the set of DAs for language $L_i$. Pooling together all of these labels into a single set  $\mathbb{T}=\bigcup\limits_{i=1}^M \mathbb{T}_{L_i}$ enables to train a multi-lingual classifier that assigns to any input text in any language $L_i \in \mathbb{L}$ a single label from $\mathbb{T}$.
Such a model is able to recognize DAs in arbitrarily many languages but 
it is necessary to retrain the model when a new language is added.

\subsection{Cross-lingual Model}
\label{sec:clm}
The cross-lingual model relies on a semantic space transformation.
It is indeed possible to transform the lexical semantic space of any language so that word representations of similar concepts in different languages are close.
Based on our previous work~\cite{martinek2018space} we chose the canonical correlation analysis (CCA) \cite{brychcin2018linear} method.
It is a technique for multivariate data analysis and dimensionality reduction, which quantifies the linear associations between a pair of random vectors.
It can be used for a transformation of one semantic space to another.

The DA recognition model is trained on a single pivot language.
The test examples from any language are then projected into the target pivot language. 
It thus allows classifying DAs in any language from within the transformed semantic space. 
Retraining the DA recognition model is not necessary when a new language is considered.

\subsection{DA Representation}
Word2vec embeddings are used to encode word semantics. 
For the cross-lingual scenario, we create a vocabulary $V$ of the $\vert V\vert$ most frequent words in the pivot language used for training. 
In the multi-lingual case the vocabulary is shared and consists of the union of the vocabularies of all available languages.

In order to benefit from parallel GPU processing, the input texts have a constant length $W$. Therefore, utterances that are longer than $W$ words are truncated, while utterances with less words are padded.

The input to each proposed neural network model is either a sequence of $W$ vocabulary indexes, when the word embedding matrix is considered as part of the model's parameters;
or directly a sequence of $W$ embedding vectors when these embeddings are considered as constant.

The advantage of the former input is the possibility to fine-tune the word vectors, while the latter option 
allows us to use the transformed semantic spaces seamlessly.

\subsection{Neural Network Topologies}

\subsubsection{Convolutional Neural Networks}
\label{sec:cnn_arch}
\begin{figure}[!ht]
\hspace{-0.9cm}
\includegraphics[width=0.43\textwidth, angle=90]{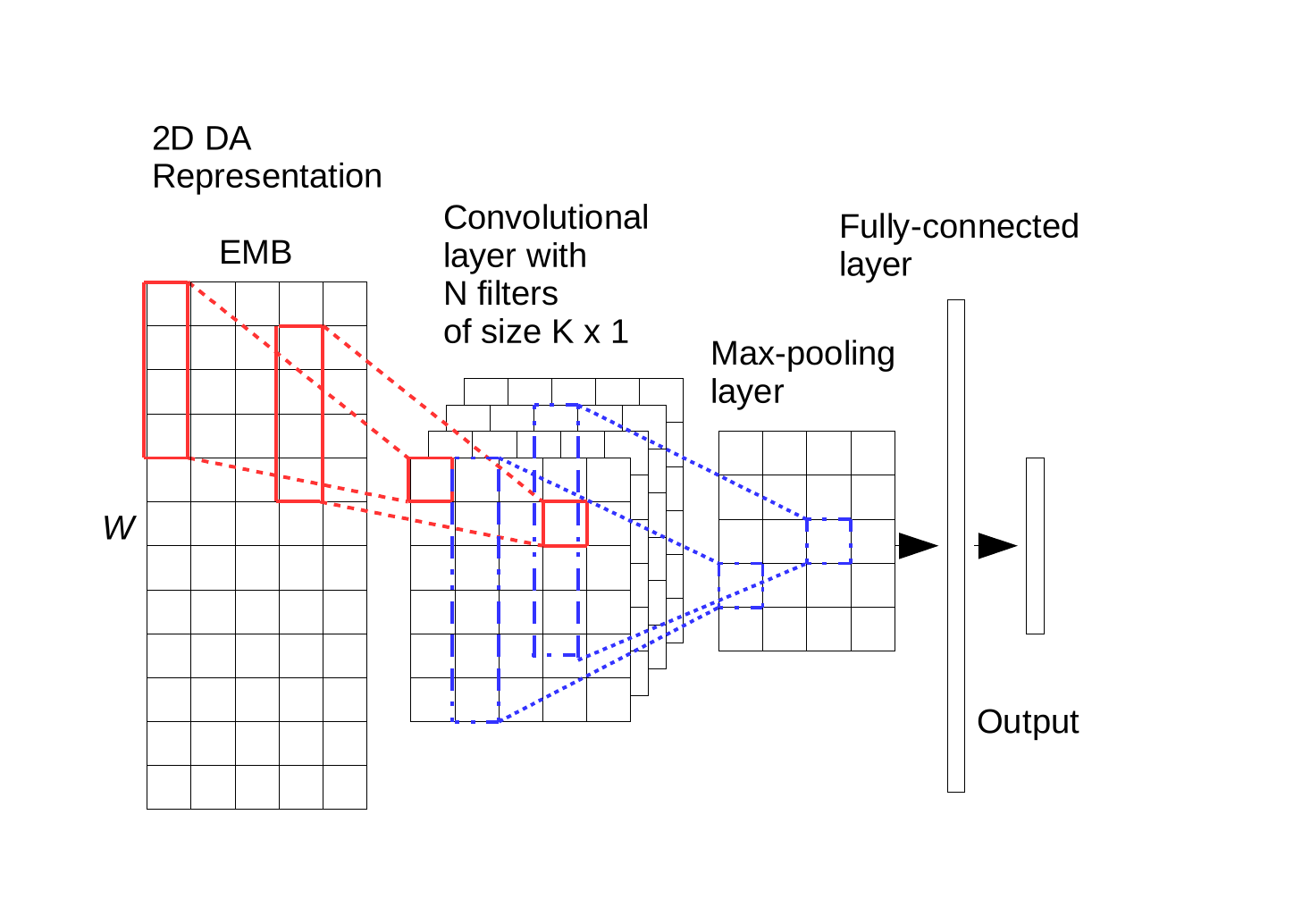}
\caption{CNN$_1$ architecture}
\label{fig:cnn_architecture}
\end{figure}

We use two CNN networks with different configurations. 
The first one is the model presented in~\cite{martinek2018space} where it was used for document classification. 
We have modified the size of the convolutional kernels to adapt them to the dialogue acts domain.
In such a domain, we usually work with much shorter inputs so we use a smaller kernel -- $(4, 1)$ as shown in Figure~\ref{fig:cnn_architecture}.

We use 40 convolutional kernels with {\it relu} activation.
A final fully-connected layer after the convolutional one consists of 256 neurons, which are further concatenated with the previous vector when we consider the history (the previous DA).
We use categorical cross-entropy as a loss function and softmax activation in the output layer.
We refer to this architecture as CNN$_1$.

The second configuration follows Kim~\cite{kim2014convolutional}. 
It uses three sizes of convolutional kernels --  $(3, EMB)$,  $(4, EMB)$ and  $(5, EMB)$ where $EMB$ is the embedding dimensionality. 
100 kernels of each size are computed simultaneously and their outputs are merged and fed into a fully connected layer. 
The final layer is the same as in the previous case.
We refer to this model as CNN$_2$.

\subsubsection{Bidirectional Long Short-Term Memory}
\begin{figure}[!ht]
\centering
\includegraphics[width=0.45\textwidth]{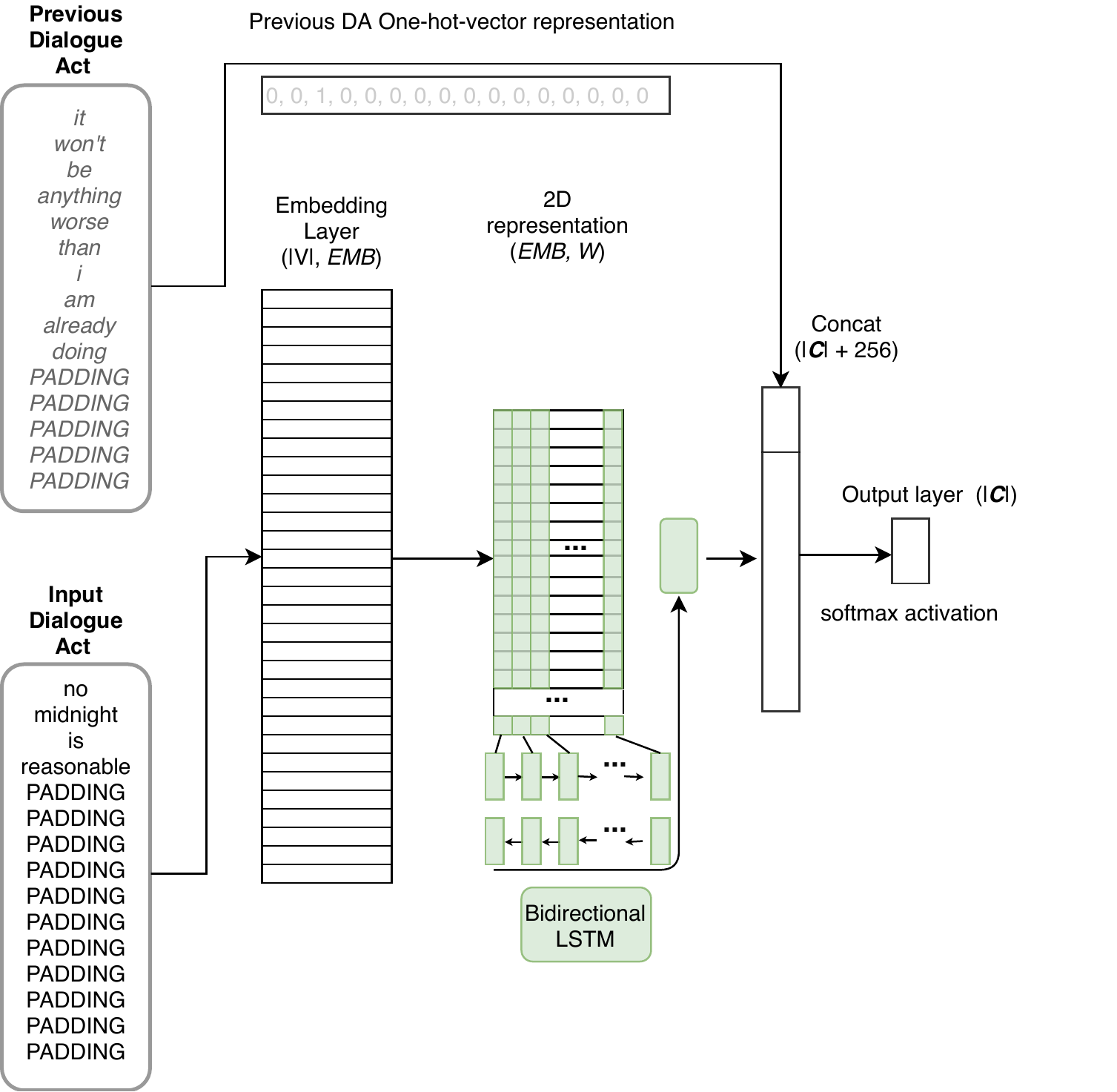}
\caption{Bidirectional LSTM architecture}
\label{fig:bilstmarchitecture}
\end{figure}

The second approach exploits a Bidirectional LSTM layer. 
The representation of the input and the embedding layer are the same as for the CNNs.
The core of this model is the Bi-LSTM layer with 100 units (i.e. 200 units in total for both directions).
The word embedding representation of the input (with $15 \times 300$ size) is fed into the Bi-LSTM layer, which outputs a single vector of 200 dimensions.
This vector is then concatenated with the predicted dialogue act class of the previous sentence encoded in the form of a one-hot-vector.
If the dialogue act has no history (e.g. initial dialogue act), a zero vector is used. 
The output layer has a softmax activation function. 
Figure~\ref{fig:bilstmarchitecture} shows this model's architecture. This model is trained using teacher forcing, while decoding is greedy.

\section{Experiments}

\subsection{Multi-lingual Verbmobil Corpus}
This corpus~\cite{alexandersson1998dialogue} was created within the Verbmobil project the goal of which was the development of a mobile application for translation of spontaneous dialogues.

It is composed of English, German as well as Japanese dialogues, however our version downloaded from LDC\footnote{\url{https://www.ldc.upenn.edu/}} contains only English and German utterances annotated with DA labels.
Therefore, we evaluate the proposed approaches on English and German.
Statistical information about this corpus is depicted in Table~\ref{table:dataset_info}.

\begin{table}[!ht]
\caption{Corpus statistical information}
\vspace{-8pt}
\begin{tabular}{l|cc|cc}
\hline
 & \multicolumn{2}{c}{English} & \multicolumn{2}{|c}{German}\\
 unit & Training & Testing & Training & Testing \\ \hline
dialogue \# & 6 485  & 940 & 15 513 & 622 \\
DA \#   & 9 599  & 1 420 & 32 269 & 1 460 \\
word \# & 79 506 & 11 086 & 297 089 & 14 819 \\
\hline
\end{tabular}
\label{table:dataset_info}
\end{table}

This dataset is annotated with 42 dialogue acts, which are grouped into the 16 following classes: \textit{feedback}, \textit{greet}, \textit{inform}, \textit{suggest}, \textit{init}, \textit{close}, \textit{request}, \textit{deliberate}, \textit{bye}, \textit{commit}, \textit{thank}, \textit{politeness\_formula}, \textit{backchannel}, \textit{introduce}, \textit{defer} and \textit{offer}.

\begin{figure}[!ht]
\centering
\includegraphics[width=0.45\textwidth]{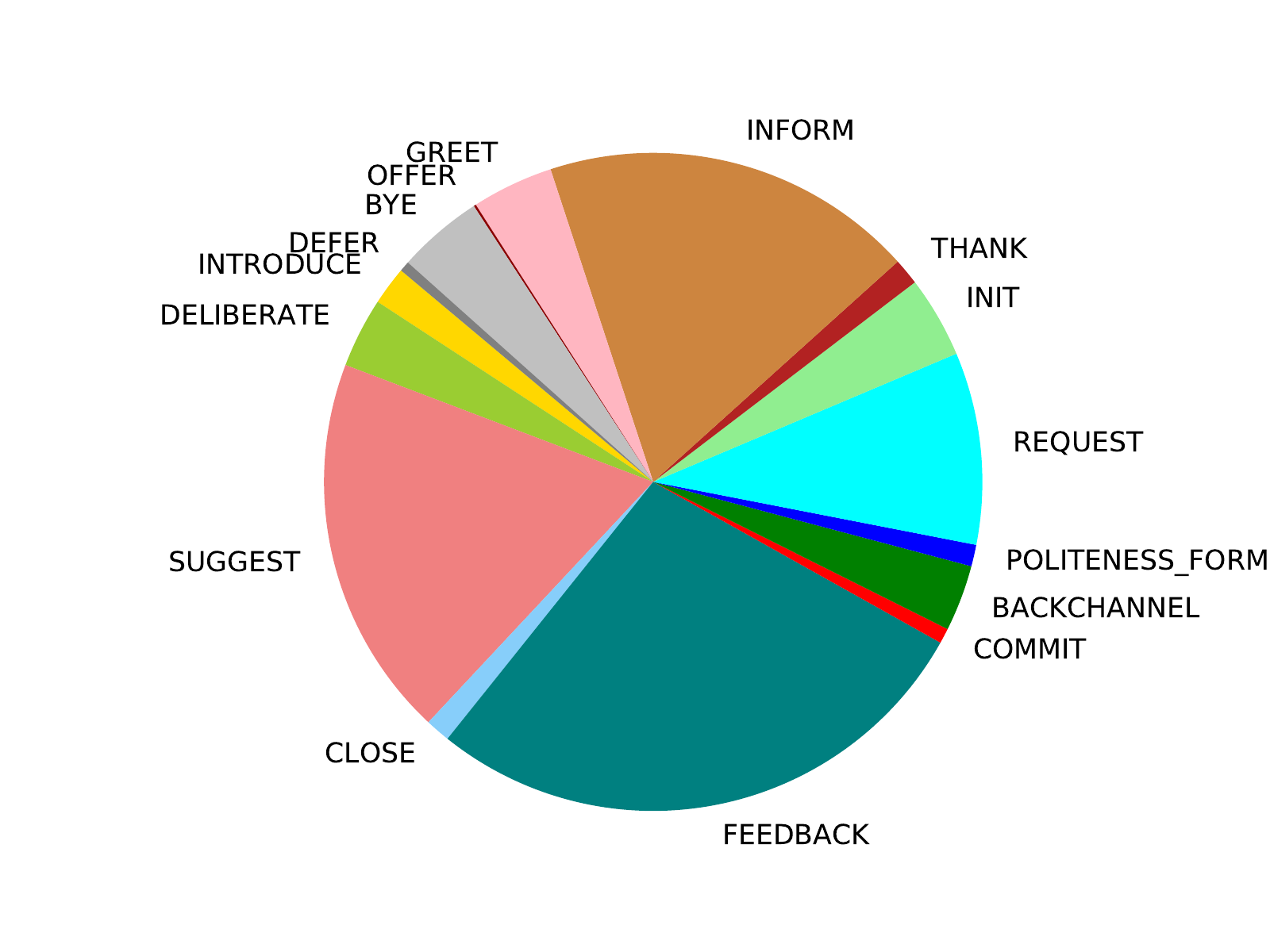}
\caption{Distribution of tags in the German training corpus part}
\label{fig:classes_count}
\end{figure}

The corpus is very unbalanced. 
In both languages, there are four dominant DAs (namely \textit{feedback}, \textit{suggest}, \textit{inform}, \textit{request}) which represent almost 80\% of the corpus size.
Figure~\ref{fig:classes_count} shows the DA distribution in the German training part.
A similar distribution is obtained on the English corpus.

\subsection{Experimental Set-up}
We use word2vec vectors trained on the English and German Wikipedia to initialize the word embeddings in our models.
Sentences are truncated or padded to 15 words in all experiments.
The vocabulary size is set to 10,000.

We evaluate all models with and without information from the dialogue history, which consists of the dialogue act that has been predicted at the previous sentence.

Although most related works use the accuracy (Acc) measure, we further compute the F1 score (macro), because the corpus is unbalanced and therefore the F1 score is more relevant.
We run all experiments 10 times and the results are averaged.

\subsection{Multi-lingual Model Results}
This series of experiments shows results of the multi-lingual model presented in Section~\ref{sec:mlm}.
Table~\ref{table:mlr_static} reports the performance of the models with static word2vec embeddings while Table~\ref{table:mlr_trainable} presents the results with fine-tuned embeddings.
These tables show that, generally, fine-tuning word2vec embeddings does not bring any improvement for DA recognition.
The relatively high differences between the accuracy and F1 score are caused by the significant corpus unbalances.
Another interesting observation is that the dialogue history helps for DA recognition in all but a few cases and that the best neural classifier is the Bi-LSTM network.

\begin{table*}[ht!]
\centering
{\footnotesize
\caption{Multi-lingual DA recognition with static word2vec embeddings}
\vspace{-8pt}
\begin{tabular}{cc|cccc|cccc|cccc}
\hline
& & \multicolumn{4}{c|}{\textbf{CNN$_1$}} & \multicolumn{4}{c|}{\textbf{CNN$_2$}} & 
\multicolumn{4}{c}{\textbf{Bi-LSTM}} \\ 
& & \multicolumn{2}{c}{\textbf{With History}} & \multicolumn{2}{c|}{\textbf{No History}} & \multicolumn{2}{c}{\textbf{With History}} & \multicolumn{2}{c|}{\textbf{No History}} & \multicolumn{2}{c}{\textbf{With History}} & \multicolumn{2}{c}{\textbf{No History}} \\ 
Train & Test & Acc & F1 & Acc & F1 & Acc & F1 & Acc & F1 & Acc & F1 & Acc & F1 \\ \hline
en    & en   & 72.1            & 58.9            & 72.2           & 58.4           & 74.5            & \textbf{65.8}            & 74.3           & 65.1           & \textbf{74.9}            & 60.5            & 74.1           & 59.9           \\
de    & de   & 72.5            & \textbf{60.8}            & 71.8           & 58.2           & 71.9            & 57.5            & 70.8           & 56.6           & \textbf{74.3}           & 59.3            & 73.6           & 59.4           \\
en+de & de   & 72.0              & 59.9            & 71.2           & 57.7           & 70.9            & 57.5            & 71.1           & 54.6           & \textbf{74.3}            & \textbf{61.7}            & 73.2           & 60.6           \\
en+de & en   & 70.3            & 55.1            & 70.0             & 55.5           & 71.4            & 57.1            & 70.7           & \textbf{58.6}           & \textbf{72.8}            & 58.5            & 72.6           & 57.4           \\ \hline
\end{tabular}
\label{table:mlr_static}
}
\end{table*}

\begin{table*}[ht!]
\centering
{\footnotesize
\caption{Multi-lingual DA recognition with fine-tuned word2vec embeddings}
\vspace{-8pt}
\begin{tabular}{cc|cccc|cccc|cccc}
\hline
& & \multicolumn{4}{c|}{\textbf{CNN$_1$}} & \multicolumn{4}{c|}{\textbf{CNN$_2$}} & 
\multicolumn{4}{c}{\textbf{Bi-LSTM}} \\

& & \multicolumn{2}{c}{\textbf{With History}} & \multicolumn{2}{c|}{\textbf{No History}} & \multicolumn{2}{c}{\textbf{With History}} & \multicolumn{2}{c|}{\textbf{No History}} & \multicolumn{2}{c}{\textbf{With History}} & \multicolumn{2}{c}{\textbf{No History}} \\ 
Train & Test & Acc & F1 & Acc & F1 & Acc & F1 & Acc & F1 & Acc & F1 & Acc & F1 \\ \hline
en    & en   & 72.2            & \textbf{69.2}            & 72.2           & 68.4           & \textbf{73.7}            & 68.4            & 72.1           & 59.1           & 73.5            & 67.2            & 72.7           & 67.2           \\ 
de    & de   & 72.7            & 59.2            & 71.7           & 57.7           & 72.1            & 59.1            & 72.6           & \textbf{60.4}           & \textbf{74.9}            & 57.2            & 74.3           & 59.1           \\ 
en+de & de   & 71.8            & \textbf{60.8}            & 70.8           & 58.9           & 70.8            & 58.4            & 71.7           & 58.8           & \textbf{72.7}            & 58.2            & 71.4           & 58.3           \\ 
en+de & en   & \textbf{69.2}            & 61.2            & 68.6           & 58.6           & 69.6            & 61.2            & 68.7           & 60.2           & 68.5            & 60.1            & \textbf{69.2}           & \textbf{63.1 }          \\ \hline
\end{tabular}
\label{table:mlr_trainable}
}
\end{table*}

\begin{table*}[ht!]
\centering
{\footnotesize
\caption{Cross-lingual DA recognition based on CCA transformation}
\vspace{-8pt}
\begin{tabular}{cc|cccc|cccc|ccccc}
\hline
& & \multicolumn{4}{c|}{\textbf{CNN$_1$}} & \multicolumn{4}{c|}{\textbf{CNN$_2$}} & 
\multicolumn{4}{c}{\textbf{Bi-LSTM}} \\
& & \multicolumn{2}{c}{\textbf{With History}} & \multicolumn{2}{c|}{\textbf{No History}} & \multicolumn{2}{c}{\textbf{With History}} & \multicolumn{2}{c|}{\textbf{No History}} & \multicolumn{2}{c}{\textbf{With History}} & \multicolumn{2}{c}{\textbf{No History}} \\
Train & Test & Acc & F1 & Acc & F1 & Acc & F1 & Acc & F1 & Acc & F1 & Acc & F1 \\ \hline
en    & de   & 30.7            & 11.9            & \textbf{34.3}           & 13.9           & 31.5            & 14.5            & 31.2            & 15.4          & 34.0              & 16.4            & 34.0              & \textbf{17.0}  \\
de    & en   & 55.1            & 26.4            & 54.4           & 25.5           & 53.9            & 28.3            & 53.0              & 27.4          & \textbf{58.6}            & \textbf{37.1}            & 57.5            & 33.7          \\ \hline
\end{tabular}
\label{table:slc}
}
\end{table*}

\subsection{Cross-lingual Model Results}
Table~\ref{table:slc} shows the results of the cross-lingual model presented in Section~\ref{sec:clm}.
The scores of the cross-lingual model are significantly lower than the scores of the multi-lingual methods reported in Tables~\ref{table:mlr_static} and \ref{table:mlr_trainable}.
The best reported accuracy is obtained by the Bi-LSTM network and it is close to 60\% when we use the German part of the corpus for training (pivot language) and the English dataset for testing.
Low F1 score occurred because of the poor results of infrequent DAs, which do not impact much the accuracy values.
The lower results for the English $\rightarrow$ German direction can be explained by the significantly smaller corpus size for training.
This table further shows that dialogue history slightly helps for DA recognition and that the Bi-LSTM significantly outperforms the two other CNN models.

\section{Comparison with Related Work}
Table~\ref{tab:sota} compares the performances of the proposed models with several state-of-the-art systems.

\begin{table}[!htb]
{\footnotesize
\caption{\label{tab:sota} Comparison with the state of the art [accuracy in \%].}
\vspace{-8pt}
\begin{center}
\vspace{-8pt}
\begin{tabular}{lc}
\hline {\bf ~Method}~& Acc \% \\
\hline
n-grams + complex features~\cite{reithinger1997dialogue} & 74.7 \\ 
TBL + complex features~\cite{samuel1998dialogue} & 71.2 \\
ME + BoW features & 49.1 \\
LSTM + w2vec features~\cite{cerisara2018effects} & 74.0\\ 
\hline
~CNN + w2vec features (proposed)~     & 74.5 \\
~Bi-LSTM + w2vec features (proposed)~ & {\bf 74.9} \\
\hline
\end{tabular}
\end{center}
}
\end{table}

We only consider in these experiments our mono-lingual English models, because we have not found any cross- or multi-lingual results in the literature about dialog act recognition to compare with.
First, we report the results of traditional feature-engineering methods, which combine a rich set of handcrafted features with dialogue history using Bayesian n-gram~\cite{reithinger1997dialogue} or TBL classifier~\cite{samuel1998dialogue}.
These methods have obtained the best score on the Verbmobil corpus so far.

We further implemented another baseline that uses a maximum entropy (ME) classifier with simple bag of words (BoW) features.
Then, we show the results of our previous LSTM system~\cite{cerisara2018effects}, which uses only simple word level features and word tokens from the previous dialogue act.
The results of our approaches are presented in the last two lines of this table.

Although we have done our best to replicate the same experimental set-up as in the related works, some doubts subsist, because the training/testing splits are not available. 
Therefore, the reported results of the first two methods may not be precisely compared with the others.
However, we can still conclude that the performance of our methods is comparable with the state of the art.

\section{Conclusions}
In this paper, we have proposed and compared several methods for multi-lingual and cross-lingual DA recognition based on deep neural networks.
The first approach builds one general model that is trained on the embeddings from all available languages, while
the second one trains the model only on one pivot language and cross-lingual projection is achieved by the CCA transform method.
We have compared and evaluated two different CNN configurations and one Bi-LSTM on the Verbmobil corpus with English and German DAs.

We have shown that the multi-lingual model significantly outperforms the cross-lingual approach.
Another advantage of the multi-lingual model is that it does not need language detection.  
However, the multi-lingual model is less flexible and may not scale easily to many languages, because retraining is necessary when adding new languages.
We have further shown that fine-tuning of word2vec embeddings does not bring any improvement in Verbmobil.
We have confirmed that the dialogue history is beneficial for DA recognition in almost all cases and that the best neural classifier is the Bi-LSTM network.
We have also compared our approaches with several state-of-the-art methods in mono-lingual scenario and concluded that the performance of our methods is comparable with the state of the art.

The generic approaches depicted in this work may be improved in many ways, e.g., by exploiting contextual word embeddings, transformer-based encoders and by annotating more languages. In the short term, it would also be interesting to improve the multi-lingual model by better handling words that have the same form across languages. 

\section{Acknowledgements}
This work has been partly supported by Cross-border Cooperation Program Czech Republic - Free State of Bavaria ETS Objective 2014-2020 (project no. 211) and by Grant No. SGS-2019-018 Processing of heterogeneous data and its specialized applications.

\bibliographystyle{IEEEtran}

\bibliography{mybib}

\end{document}